\documentclass[letterpaper, 10 pt, conference]{ieeeconf}  

\IEEEoverridecommandlockouts                              

\usepackage{graphics} 
\usepackage{graphicx}
\usepackage{amsmath} 
\usepackage{amssymb}  
\usepackage{color}
\usepackage{siunitx}
\usepackage{ifthen}
\newboolean{ArXiVversion}
\setboolean{ArXiVversion}{true}

\DeclareMathOperator*{\argmin}{arg\,min}

\newcommand{\norm}[1]{\left\lVert#1\right\rVert}

\definecolor{myBlue}{RGB}{47,139,235}

\usepackage{fancyhdr}

\ifthenelse{\boolean{ArXiVversion}}{
    
    \fancypagestyle{firstpage}{
        \fancyhf{}

        \fancyhead[C]{}
    
    	\fancyfoot[C]{\strut \copyright~This work has been submitted to the IEEE for possible publication. Copyright may be transferred without notice, after which this version may no longer be accessible.}
    }
}{}

\title{\LARGE \bf
Reacting on human stubbornness in human-machine trajectory planning
}

\author{Julian Schneider, Niels Straky, Simon Meyer, Balint Varga and Sören Hohmann
\thanks{All authors are with Faculty of Electrical Engineering, Institute of Control Systems (IRS),
        Karlsruhe of Technology (KIT), 76131 Karlsruhe, Germany
        {\tt\small julian.schneider@kit.edu}}%
}

\begin{document}

\maketitle
\ifthenelse{\boolean{ArXiVversion}}{
    \thispagestyle{firstpage}
    
}{}
\pagestyle{empty}

\begin{abstract}

In this paper, a method for a cooperative trajectory planning between a human and an automation is extended by a behavioral model of the human. This model can characterize the stubbornness of the human, which measures how strong the human adheres to his preferred trajectory. Accordingly, a static model is introduced indicating a link between the force in haptically coupled human-robot interactions and humans's stubbornness. The introduced stubbornness parameter enables an application-independent reaction of the automation for the cooperative trajectory planning. Simulation results in the context of human-machine cooperation in a care application show that the proposed behavioral model can quantitatively estimate the stubbornness of the interacting human, enabling a more targeted adaptation of the automation to the human behavior.

\end{abstract}

\section{INTRODUCTION} \label{sec:introduction}

With the advent of Industry 4.0, it's conceivable that Care 4.0 could be next \cite{ahsan_industry_2022}. There exists considerable unexplored potential in robotic systems within the caregiving area \cite{kehl_robotics_2018}. The support of intelligent systems could enable people in need of care longer independent living, possibly in their own homes \cite{draper_ethical_2017}. Robotic systems also have the potential to lessen the physical workload of caregiving personnel or healthcare staff \cite{draper_ethical_2017}. Besides logistical operations like administering medication or meal provision \cite{bloss_mobile_2011}, caregiving inherently involves physical human contact \cite{burgess_nurses_2023,abdulazeem_human_2023}. This physical human contact is already evident in individual robotic care applications like patient accompaniment to medical examination rooms \cite{schneider_negotiation-based_2022}, but is mainly present in other robotic application areas like manufacturing \cite{cherubini_collaborative_2016}, imitation learning \cite{lee_mimetic_2010}, or indirect physical coupling via a jointly held object \cite{abdulazeem_human_2023,kosuge_mobile_2000}. These examples showcase interactive human-robot systems that present intriguing possibilities for the use of human-machine cooperation in caregiving context.

The literature describes models of human-machine cooperation frameworks that categorize such interactions into three or four distinct levels \cite{abbink_topology_2018,pacaux-lemoine_layers_2019,rothfus_study_2020}. In \cite{rothfus_study_2020} cooperative tasks are divided into \textit{decomposition level}, \textit{decision level}, \textit{trajectory level} and \textit{action level}. At the action level, many studies focus on modeling these interactions \cite{inga_evaluating_2018,abbink_importance_2012, mars_modelling_2017} and developing controllers for human-machine interaction \cite{abbink_neuromuscular_2010, teodorescu_assistme_2022}. On the other hand, there is less work on cooperation at the trajectory level within human-machine cooperation research, where it is the goal to agree on a common reference trajectory. Examples of works on this level are  only from the automotive domain \cite{boink_understanding_2014, huang_human-machine_2022,benloucif_cooperative_2019}. Nevertheless, reaching this agreement is vital for its application in elderly care. Therefore, a method for the cooperative agreement on a common trajectory between a human and an automation was presented in \cite{schneider_negotiation-based_2022} for a haptically coupled human-automation system, in which the cooperative agreement on a common trajectory is found based on negotiation. The behavior of the human for this negotiation has been described with a simple heuristic evaluation. According to \cite{inga_evaluating_2018}, however, the description of human behavior plays a central role in the area of shared control, as it is an important evaluation criterion for deciding how automation should adapt to humans. In order to have a more intuitive and personalizable negotiation between a human and a machine, a general human behavior model is necessary that is able to characterize the human. This enables a targeted response of the automation based on the estimated \textit{stubbornness} of the human. Such a behavioral model is introduced in this paper, which describes people’s behavior concerning their adherence to their desired trajectory, i.\,e. their stubbornness, or, to put it another way, their willingness to deviate from their desired trajectory, i.\,e. their concession. In this paper, stubbornness is considered to be the opposite of \textit{concession}.

The contribution involves expanding an existing framework to include a generalized parameter that encapsulates human stubbornness. This modification enables the automation to tailor its response to the human actions. By introducing this generalized parameter, the proposed framework can be extended to other applications or other communication flows between the human and the automation.

The main challenge is that the desired reference trajectory of the human is not known for the automation which is an existing problem in human-machine interaction in unstructured environments, see e.\,g. \cite{varga_2020, varga_2023}. This challenge is addressed in this paper with the assumption that the human communicates his stubbornness via a measurable interaction force from which the automation estimates the stubbornness of the humans.

The paper is structured as follows. Section \ref{sec:related_work} presents the related works on cooperative trajectory planning and on modeling human behavior of stubbornness in motion scenarios. Section \ref{sec:collab_trajectory_planning} presents the concept of the cooperative trajectory planning that is used (Subsection \ref{subsec:introduction_coop_trajec_planning}) and how it is further developed by the human stubbornness model (Subsection \ref{subsec:further-development}). Section \ref{sec:simulation_results} describes a simulation scenario and shows simulative results and Section \ref{sec:discussion} classifies them.

\begin{figure}[t]
   \centering
   \includegraphics[width=\columnwidth]{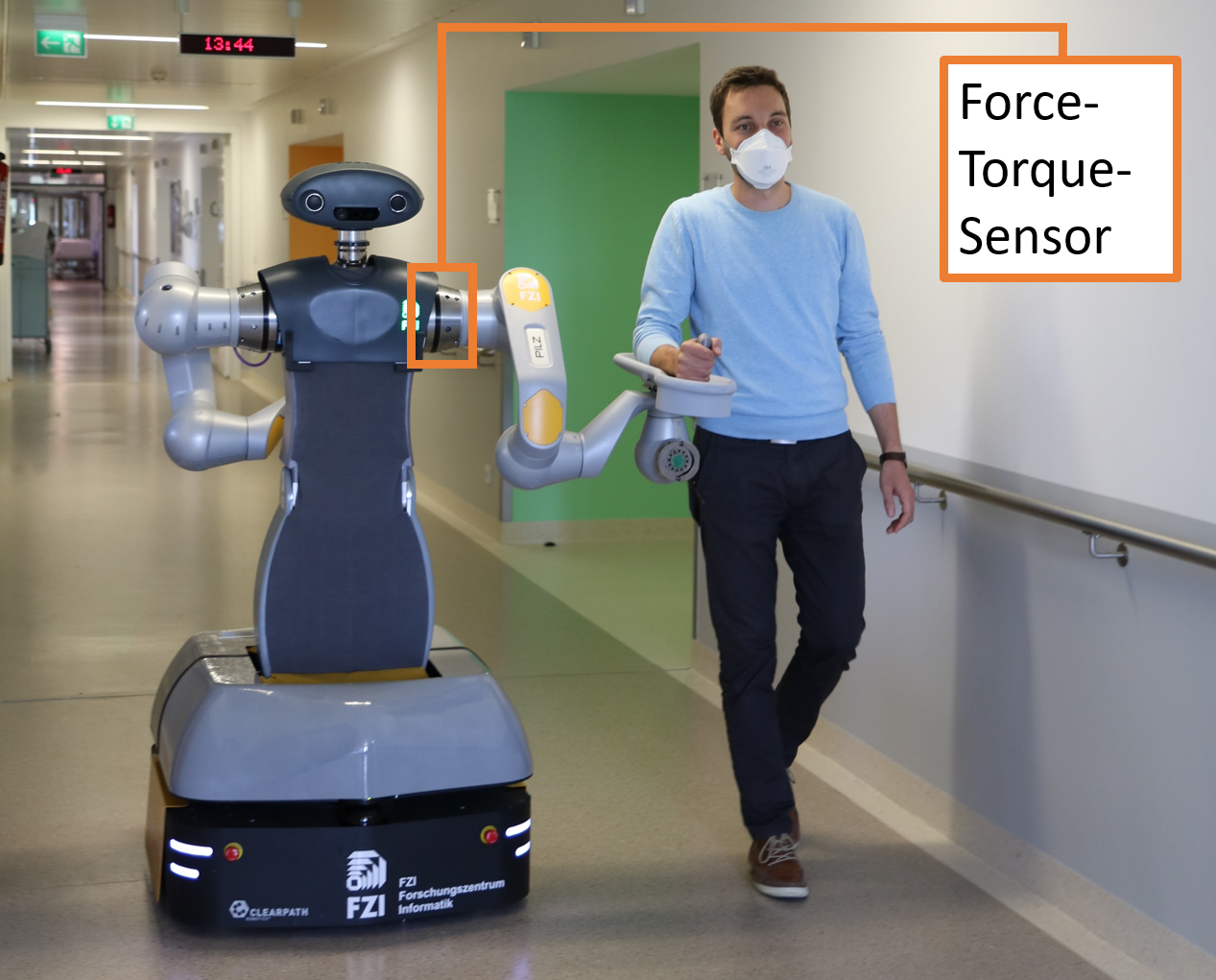}
   \caption{The application under consideration: Accompanying a patient in a hospital.}
   \label{fig:application}
\end{figure}

\section{RELATED WORK} \label{sec:related_work}
\subsection{Cooperative Trajectory Planning}
This subsection provides an overview of the existing approaches to cooperative trajectory planning in the context of human-machine cooperation research as well as a differentiation from our approach.

Cooperative trajectory planning is first addressed in the context of human-machine cooperation or shared control in \cite{boink_understanding_2014} using the application of automated driving. Here, an automation system attempts to reproduce human driving behavior from test drives by identifying a control law from driving data. The agreement on a common trajectory is therefore such that the desired human trajectory serves as a reference for the cooperative system. From a leader-follower perspective, the human is the leader in this case. \cite{benloucif_cooperative_2019} is another work that deals with cooperative trajectory planning in the context of automated driving. Here, the human's target parameters are estimated and used to form a trajectory. Again, the human is the leader. However, this is only taken into account in a weighted manner depending on an attention parameter of the automation. \cite{huang_human-machine_2022} also deals with cooperative trajectory planning in the context of automated driving. Here, the automation provides a superimposed actuating moment for the human when unsafe human driving behavior is detected. All three works presented ultimately implement a leader-follower system structure, with either the human or the automation acting as the leader.

Another cooperative trajectory planning approach is presented by \cite{schneider_negotiation-based_2022}, in which the movement requests of the human and the automation are given equal consideration. A joint trajectory is negotiated between the human and the automation using a reciprocal tit-for-tat approach. The automation estimates the human's desired movement using an interaction force measured by a force-torque sensor.

\subsection{Modeling human stubbornness in human-machine cooperation} \label{subsec:hidden_states}
The authors assert that there are no studies or existing literature that have determined the effect of human stubbornness on maintaining a particular trajectory. Although \cite{wang_2021} deduces the necessity of mutual adaptation in human-machine cooperation from the general characteristic of humans to be stubborn, he does not provide any model for human stubbornness. In \cite{pandya_towards_2023}, human stubbornness is considered in the context of human-machine cooperation, but only in the form of a fixed state value in the state space.  In contrast, however, a continuous stubbornness model is required for the approach here. This means there is no stubbornness model yet validated in the literature for the present application. Instead of formulating a new, very application-specific model, this paper formulates human stubbornness as a hidden state $x$ that can be inferred via observation, like it is the idea in hidden Markov models. In contrast to this kind of model, which is stochastic and discrete in value, the model here shall be deterministic and continuous in value. The focus here is on the observation model ${\boldsymbol{o} = f(\boldsymbol{x}_\text{stubb,H})}$, i.\,e. the model that produces an output $\boldsymbol{o}$ from the stubbornness $\boldsymbol{x}_\text{stubb,H}$ of the human. This paper aims to find the inverse function for this, i.\,e. $\boldsymbol{x}_\text{stubb,H} = f^{-1}(\boldsymbol{o})$.

\section{CONCEPT OF THE COOPERATIVE TRAJECTORY PLANNING APPROACH} \label{sec:collab_trajectory_planning}
\subsection{Introduction to the cooperative trajectory planning approach used}\label{subsec:introduction_coop_trajec_planning}
\begin{figure*}[t]
   \centering
   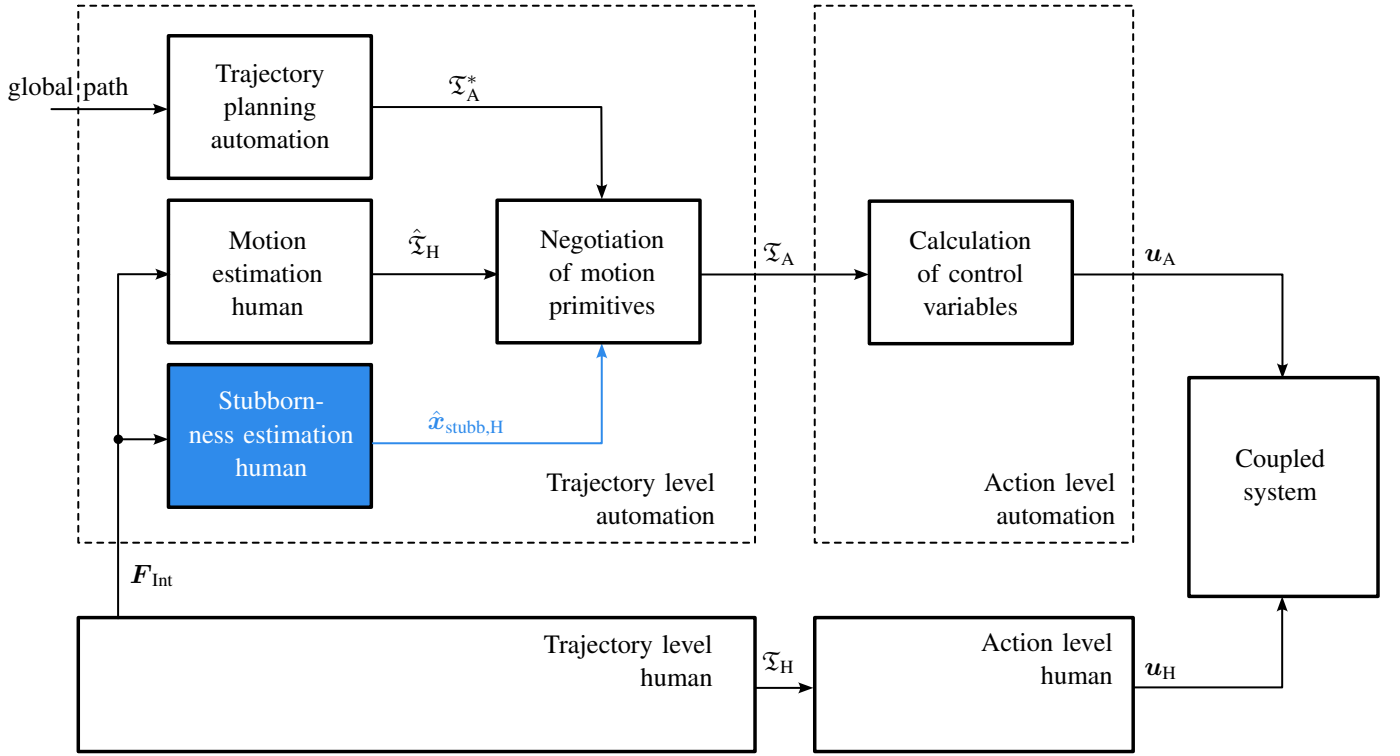
   \caption{Flowchart of the further developed automation part of cooperative trajectory planning by inserting an additional subsystem for estimating the human stubbornness that results in the parameter $\hat{\boldsymbol{x}}_\text{stubb,H}$ (marked in blue), which represents human's stubbornness.}
   \label{fig:ros2_nodes}
\end{figure*}
In this subsection, the emancipated, cooperative trajectory planning approach from the work of \cite{schneider_negotiation-based_2022} is presented in more detail, which is further developed in this paper. The core idea of this cooperative trajectory planner is that the automation suggests a trajectory that is as close as possible to the human's requests or, in the case of deviating movement requests, accommodates the human in a targeted manner and determines an agreement with the human's movement requests. 
The first input variable for the automation is a global path, which results from an upstream global path planning subsystem (see Fig. \ref{fig:application}). The trajectory planning of the automation uses this to plan a local trajectory in the form of a movement primitive $$\boldsymbol{p}(t)=[v(t), \omega(t)]^T.$$ A circular arc is selected as the movement primitive in this work that is defined by a linear speed $v(t)$ and an angular velocity $\omega(t)$. By adding a time horizon $\tau$, $\boldsymbol{p}$ thus describes a trajectory $\mathfrak{T}=[\boldsymbol{p}, \tau]^T$ within the time horizon $\tau$, where $v(t)$ and $\omega(t)$ are assumed to be constant within $\tau$, i.e. $t \in [t_0, t_0+\tau]$ where $t_0$ describes the current point in time. Since $v(t)$ and $\omega(t)$ are chosen to be constant, the time dependency $v(t)$ and $\omega(t)$ is omitted in the following and only $v$ and $\omega$ is written. The optimal movement primitive for the automation is determined using an optimization-based procedure. It results from
\begin{equation}
   \boldsymbol{p}_\text{A}^*=\argmin_{v,\omega}{J_\text{MP}}
\end{equation}
where
\begin{align}
   J_\text{MP} &= w_1 \cdot d(v,\omega)^2 + w_2 \cdot v^2+w_3 \cdot \omega^2,\\
   v&\in[v_\text{lb}, v_\text{ub}], \nonumber\\
   \omega&\in[\omega_\text{lb}, \omega_\text{ub}].\nonumber
\end{align}
The indices lb and ub represent the lower and the upper bounds of the optimization variables $v$ and $\omega$. The terms in $J$ penalize the distance between $t_0$ and the global path of the automation, which depends on the current and the necessary kinetic energy, which is proportional to $v^2$ and $\omega^2$. The $w_i$ are weighting parameters.

The motion primitive of the human is estimated in the step \textit{motion estimation human} (Fig. \ref{fig:ros2_nodes}) from the interaction force $\boldsymbol{F}_\text{Int}$ measured in the force torque sensor whereby $F_\text{Int}$ is separated into its two planar components $\boldsymbol{F}_\text{Int} = [F_\text{Int,x}, F_\text{Int,y}]^T$, where $F_\text{Int,x}$ describes the longitudinal component of $\boldsymbol{F}_\text{Int}$ and $F_\text{Int,x}$ is the lateral component. These components are interpreted as the acceleration request $\Delta v$ (longitudinal movement) and $\Delta \omega$ (lateral movement). Here, the relationship 
\begin{align*}
   \Delta v &\propto F_\text{Int,x},\\
   \Delta \omega &\propto F_\text{Int,y}
\end{align*}
for the human's request for movement is assumed where the component $F_\text{Int,x}$ indicates an acceleration request of the linear speed $v$ and $F_\text{Int,y}$ shows an acceleration request of the angular velocity $\omega$. The calculation formula for $\Delta \hat{v}$ is
\begin{equation} \label{eq:v_estimation_human}
   \resizebox{\columnwidth}{!}{%
   $\Delta \hat{v} = \begin{cases}
      \text{sgn}(F_\text{Int,x}) \dfrac{\min(|F_\text{Int,x}|, F_\text{Int,x,max})}{F_\text{Int,x,max}} v_\text{ub} \cdot f_v, &F_\text{Int,x} \geq F_\text{thresh,x} \\
      0, & \text{else}
      \end{cases}$
   }
\end{equation}
and for $\Delta \hat{\omega}$ analogously follows
\begin{equation} \label{eq:omega_estimation_human}
   \resizebox{\columnwidth}{!}{%
   $\Delta \hat{\omega} = \begin{cases}
      \text{sgn}(F_\text{Int,y}) \dfrac{\min(|F_\text{Int,y}|, F_\text{Int,y,max})}{F_\text{Int,y,max}} \omega_\text{ub} \cdot f_\omega, &F_\text{Int,y} \geq F_\text{thresh,y} \\
      0, & \text{else}.
      \end{cases}$
   }
\end{equation}
Here, the roof describes an estimated size. The signum function guarantees the correct sign, the min-function ensures that the considered interaction force is capped and the fraction with $F_\text{Int,x,max}$ or $F_\text{Int,y,max}$ in the denominator ensures normalization and thus a percentage of $v_\text{ub}$ and $\omega_\text{ub}$ as the desired accelerations. The parameters $f_v$, $f_\omega$ represent tuning or sensitivity parameters. This results in the total estimated motion primitive of the human via
\begin{equation} \label{eq:movement_estimation_human}
   \hat{\boldsymbol{p}}_\text{H}=\boldsymbol{p}+\Delta\hat{\boldsymbol{p}}=[v+\Delta \hat{v}, \omega + \Delta \hat{\omega}]^T.
\end{equation}
In the subsequent step \textit{Negotiation of motion primitives}, the movement desires of both actors are negotiated. This is done using a utility function $U$, which assigns a utility $U \in [0,1]$ to each possible movement primitive from the automation's point of view. The utility function $U$ is parameterized in such a way that the automation's movement request $\boldsymbol{p}_\text{A}^*$ provides the utility value 1. The movement primitive of the human provides the utility value 0:
\begin{align}
   U(v,\omega)&=\dfrac{J_\text{MP}(\hat{\boldsymbol{p}}_\text{H})-J_\text{MP}(v,\omega)}{J_\text{MP}(\hat{\boldsymbol{p}}_\text{H})-J_\text{MP}(\boldsymbol{p}_\text{A}^*)},\\
   v&\in[v_\text{lb}, v_\text{ub}],\nonumber\\
   \omega&\in[\omega_\text{lb}, \omega_\text{ub}].\nonumber
\end{align}
In the course of the negotiation, the automation does not stubbornly adhere to its movement primitive, but is prepared in each negotiation step to accept a certain loss of utility $U_\text{loss}$ and to deviate by this value from its maximum utility value up to a target utility value $U_\text{target}$. This target utility value can be interpreted as the minimum required utility in each negotiation round and is calculated as follows:
\begin{equation}\label{eq:u_target}
   U_\text{target} = U_\text{max}-U_\text{loss} = 1-U_\text{loss}.
\end{equation}
In contrast to time-based strategies from the literature \cite{rothfus_study_2020}, a reciprocal tit-for-tat-strategy is selected, i.\,e. a behavior-based offer strategy, for calculating $U_\text{loss}$. The special feature is that the automation acts in the opposite way to human behavior: if the human is stubborn, the automation reacts in a particularly compliant manner. If the human is more compliant, the automation tends to dictate the direction of movement (more stubborn automation). The calculation of $U_\text{loss}$ is based on the observation of human behavior in three steps:
\begin{enumerate}
   \item Heuristic determination of an interaction value $\Delta I_k$, which represents the concession behavior of the person in each time step $k$ from the effect on the interaction force.
   \item Summing up the $\Delta I_k$ over $l$ time steps results in the variable $M$ that is calculated via 
   \begin{equation}
      M=\sum_{j=1}^l\Delta I_j.
   \end{equation}
   \item The  utility loss $U_\text{loss}$ of the automation is then calculated as a function of $M$ according to
   \begin{equation}
      U_\text{loss}= \begin{cases}
         0,~M<0 \\
         f_\text{M}\cdot\frac{1}{(2l)^2}\cdot M^2,~M\geq 0.
      \end{cases}
   \end{equation}
\end{enumerate}  
The negotiated movement primitive is then calculated using \eqref{eq:u_target} from
\begin{equation}
   \boldsymbol{p}_\text{A}=\argmin_{v,\omega}\{|U(v,\omega)-U_\text{target}|\}.
\end{equation}
Together with the time horizon $\tau$, $\boldsymbol{p}_\text{A}$ then forms the trajectory $\mathfrak{T}_\text{A}$ executed by the automation, which corresponds to an agreement with the estimated desire $\hat{\mathfrak{T}}_\text{H}$ of the human. The trajectory $\mathfrak{T}_\text{A}$ is then converted into the control variables $\boldsymbol{u}_\text{A}$ in the step \textit{Calculation of control variables} (Fig. \ref{fig:ros2_nodes}), which represent the Action level of the human-machine cooperation level model of \cite{rothfus_study_2020}.

\subsection{Further development} \label{subsec:further-development}
The main contribution of the presented emancipated trajectory planner is the introduction of the reciprocal tit-for-tat behavior model, according to which the automation behaves in the opposite direction to the human behavior. This ensures that an agreement is always found. If the human sticks very stubbornly to his desire to move, the automation acts compliantly. The movement primitive $\boldsymbol{p}_\text{A}$ resulting from the agreement is based on the determination of a target utility $U_\text{target}$ of the automation. The more stubborn the human is, the lower this target utility is. Currently, $U_\text{target}$ is calculated using the three steps described at the end of section \ref{subsec:introduction_coop_trajec_planning}, which is based on an application-specific heuristic mentioned in step 1) via the determination of $\Delta I_k$. In order to obtain a more generalized procedure and thus to decouple the calculation of $U_\text{target}$ from $\boldsymbol{F}_\text{Int}$, $U_\text{target}$ should be determined as a function of a stubbornness of the human $\boldsymbol{x}_\text{stubb,H}$ to be determined. This has the advantage that the same negotiation-based agreement model can be used for different applications or, for example, with different communication flows to communicate the desire to move with a certain stubbornness. The system structure and the idea of the further development are shown by the blue elements in Fig. \ref{fig:ros2_nodes}. The goal is to calculate a concession $U_\text{target}$ for the automation depending on the stubbornness $\boldsymbol{x}_\text{stubb,H}$ of the human. Neither the desired trajectory nor the stubbornness of the human is known, but only the interaction force $\boldsymbol{F}_\text{Int}$ and the system behavior is measurable. Therefore, a model $\boldsymbol{x}_\text{stubb,H}(\boldsymbol{F}_\text{Int})$ is determined and after that a $U_\text{target}(\boldsymbol{x}_\text{stubb,H})$ is formed.

From the authors' point of view, there are no studies to date that estimate a stubbornness o with respect to a goal adherence from a human action. We will introduce this according to the non measurable state idea from Section \ref{subsec:hidden_states}. As a starting point, the stubbornness of the human $\boldsymbol{x}_\text{stubb,H}$ is modeled in this work as a hidden state. This state itself cannot be measured directly, but is expressed in measurable observations $\boldsymbol{o}$. The observable variables $\boldsymbol{o}$ in this work are the interaction force $\boldsymbol{F}_\text{Int}$ as well as the estimated movement request of the human $\Delta \hat{\boldsymbol{p}}_{\text{H},k-1}$ from the previous time step and the movement $\boldsymbol{p}_{\text{A,k}}$ that was considered and actually executed in the current time step by the automation. Our measurement model for human stubbornness in the present application is based on the following two assumptions:
\begin{enumerate} 
   \item An increasing force amplitude indicates an increasing dissatisfaction and therefore an increasing stubbornness and
   \item The higher the movement execution deviates from the human's desired movement, the higher is the human's stubbornness on his movement desire.
\end{enumerate}
These two aspects are translated into a calculation rule for human stubbornness $\boldsymbol{x}_\text{stubb,H} = [x_\text{stubb,H,v}, x_{\text{stubb,H,}\omega}]^T$ using the following formula divided into the longitudinal and the lateral component of the stubbornness:
%
%
%
\begin{align} 
   & x_\text{stubb,H,v} = \left| \dfrac{F_\text{Int,x}}{F_\text{Int,x,max}} \right| \nonumber\\ &\cdot \left| \dfrac{\max(|\hat{v}_{\text{H},k-1}-v_\text{ub}|, |\hat{v}_{\text{H},k-1}-v_\text{lb}|)-|\hat{v}_{\text{H},k-1}-v_{\text{A},k}|}{\max(|\hat{v}_{\text{H},k-1}-v_\text{ub}|, |\hat{v}_{\text{H},k-1}-v_\text{lb}|)}\right| \label{eq:stubbornness_v} \\
   & x_{\text{stubb,H,}\omega} = \left| \dfrac{F_\text{Int,y}}{F_\text{Int,y,max}} \right| \nonumber\\ &\cdot \left| \dfrac{\max(|\hat{\omega}_{\text{H},k-1}-\omega_\text{ub}|, |\hat{\omega}_{\text{H},k-1}-\omega_\text{lb}|)-|\hat{\omega}_{\text{H},k-1}-\omega_{\text{A},k}|}{\max(|\hat{\omega}_{\text{H},k-1}-\omega_\text{ub}|, |\hat{\omega}_{\text{H},k-1}-\omega_\text{lb}|)}\right|. \label{eq:stubbornness_omega}
\end{align} 

The first component takes into account the amplitude of the interaction force and implements the assumption that a high amplitude of the interaction force is associated with great stubbornness and, accordingly, adherence to one's own desire to move. The second factor implements the degree of consideration of the human desire to move that preceded the current one. This is achieved by the feedback of the currently executed trajectory, i.e. the unified trajectory from the current time step $k$, and the estimated human movement request from the previous time step $k-1$. This comparison of the current movement with the human request from the past is shown in the difference between the human request $\hat{v}_{H,k-1}$ and $\hat{\omega}_{H,k-1}$ (target values) and the currently executed trajectory $v_{\text{A},k}$ and $\omega_{\text{A},k}$ (actual values). This results in a control deviation from the human perspective due to the consideration of the automation's desire to move, which leads to stubbornness.

\begin{figure}[t!]
   \centering
   \includegraphics{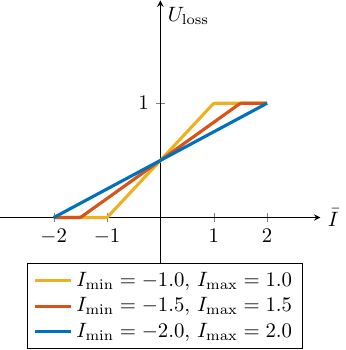}
   \caption{Curves for $U_\text{loss}$ for three different variants of parameters $\bar{I}_\text{min}$ and $\bar{I}_\text{max}$.}
   \label{fig:u_loss_varianten}
\end{figure}

With the norm of the determined stubbornness of the human $ \norm{\boldsymbol{x}_{\text{stubb},\text{H}}}$, an interaction value $I_k$ is then calculated for each time step which additionally relates the stubbornness of the human to the behavior of the automation. If the automation is completely compliant ($U_\text{loss}=1$) and the human is nevertheless very stubborn ($\norm{\boldsymbol{x}_{\text{stubb},\text{H}}}=1)$, the interaction value $I_k$ results in the maximum possible value of $I_k=2$, which describes the maximum stubbornness of the human despite compliant automation:
\begin{equation}
   I_k=-2+3\cdot \norm{\boldsymbol{x}_{\text{stubb},\text{H}}}+U_\text{loss}.
\end{equation}
To smooth the interaction values, an average value is formed over $l$ time steps
\begin{equation} \label{eq:interaction_value_average}
   \bar{I}=\dfrac{1}{l}\cdot \sum_{j=1}^l I_j
\end{equation}
and from this $U_\text{loss}$ is calculated with selectable parameters $\bar{I}_\text{min}$ and $\bar{I}_\text{max}$, which ensure an adjustable sensitivity:
\begin{equation}
   U_\text{loss}=\dfrac{\bar{I}-\bar{I}_\text{min}}{\bar{I}_\text{max}-\bar{I}_\text{min}}.
\end{equation}
Fig. \ref{fig:u_loss_varianten} shows different curves for $U_\text{loss}$ depending on the selected parameters $\bar{I}_\text{min}$ and $\bar{I}_\text{max}$. $U_\text{target}(\boldsymbol{x}_\text{stubb,H})$ is finally calculated using equation \eqref{eq:u_target}.

\section{SIMULATION} \label{sec:simulation_results}
\subsection{Simulation scenario description}
The simulation in this work models a scenario in which the human and the automation walk along a straight corridor and the human wants to avoid a puddle. The human communicates his wish to evade the puddle via the interaction force $\boldsymbol{F}_\text{Int}$. The simulation is an open-loop simulation in which the interaction force exerted by the human causes an evading maneuver by the automation. However, the following feedback of the evasive behavior on the exerted force $\boldsymbol{F}_\text{Int}$ of the person is not considered in this work. The kinematic unicycle model is used as the system model for the movement of the automation:
\begin{align} \label{eq:system-model}
   \begin{bmatrix}
      \dot{x}_\text{A}\\
      \dot{y}_\text{A}\\
      \dot{\theta}_\text{A}
      \end{bmatrix} = \begin{bmatrix}
         v_\text{A}\cdot \cos\theta_\text{A} \\
         v_\text{A}\cdot \sin\theta_\text{A} \\
         \omega_\text{A}
      \end{bmatrix}.
\end{align}
For better comprehensibility, it is assumed that the negotiation of the joint movement request only takes place in the $\omega$-dimension. It is therefore assumed that both agents walk along the corridor at a constant longitudinal speed of $v=\SI{0.8}{\metre \per \second}$ and that the human only communicates a lateral interaction force $F_\text{Int,y}$. The global path for the automation is assumed to be given and is always at $y=0$ in the lateral components. 
\subsection{Simulation results}
\begin{figure}[t!]
   \centering
   \includegraphics{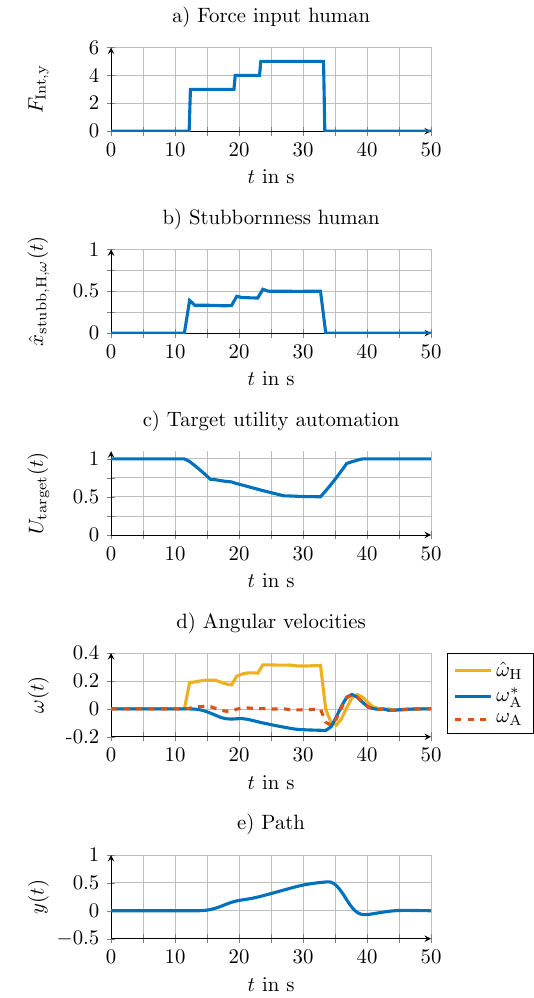}
   \caption{Plot of simulation results}
   \label{fig:sim_results}
\end{figure}
In the simulation scenario the human communicates the desire to move away to the automation via $F_\text{Int,y}$ from $t=\SI{12}{\second}$, as can be seen in Fig. \ref{fig:sim_results} a), initially with $F_\text{Int,y}=\SI{3}{\newton}$. To emphasize his request to move, he increases the force to $\SI{4}{\newton}$ at $t=\SI{19}{\second}$ and to $\SI{5}{\newton}$ after $t=\SI{23}{\second}$. From $t=\SI{33}{\second}$, the two agents have passed the obstacle and the human no longer exerts any force.

Fig. \ref{fig:sim_results} b) shows the estimation of the stubbornness $\hat{x}_{\text{stubb,H},\omega}$ of the person according to equation \eqref{eq:stubbornness_omega}. It can be seen that the estimated stubbornness $\hat{x}_{\text{stubb,H},\omega}$ is very similar to the force input from Fig. \ref{fig:sim_results} a). This results from the estimation model \eqref{eq:stubbornness_omega}, according to which the force amplitude is significantly included. In addition, overshoots can be seen in the stubbornness estimation at the times of the force leaps of $F_\text{Int,y}$. These result from the fact that the estimation after the first force leap is initially $\hat{x}_{\text{stubb,H},\omega}\approx 0.4$. However, since the desired movement $\hat{\omega}_\text{H}$ of the human and the executed movement of the system $\omega_\text{A}$ differ in the next time step, the stubbornness $\hat{x}_{\text{stubb,H},\omega}$ of the human is scaled down to $\hat{x}_{\text{stubb,H},\omega}\approx 0.33$ in the following time step according to the second term from \eqref{eq:stubbornness_omega} (human is slightly compliant if the desired movement and the executed movement differ; otherwise he would exert an increasing force). The same effects can be seen in the subsequent force leaps at $t=\SI{19}{\second}$ and $t=\SI{23}{\second}$, after which the stubbornness of the person adjusts to $\hat{x}_{\text{stubb,H},\omega}\approx0.44$ and $\hat{x}_{\text{stubb,H},\omega}\approx 0.5$ respectively.

Fig. \ref{fig:sim_results} c) shows the resulting target utility $U_\text{target}$ of the automation, i.e. the willingness of the automation to deviate from its desired trajectory. It can be seen that $U_\text{target}$ starts at the value $U_\text{target}=1$. It can also be seen that its value decreases as the stubbornness of the human increases (Fig. \ref{fig:sim_results} b)). This precisely reflects the desired reciprocal tit-for-tat behavior: the automation reacts to a non-stubborn human ($x_\text{stubb,H}=0$) in such a way that it exercises its desire to move. As the stubbornness of the human increases, the automation increasingly abandons its desire to move. After the stubbornness of the human increases from $t=\SI{19}{\second}$, the target utility value of the automation drops sharply until $t=\SI{15.5}{\second}$ and then decreases with a flatter gradient until $t=\SI{19}{\second}$. The flattening results from the selected parameter $l$, over which the interaction value $\bar{I}$ in \eqref{eq:interaction_value_average} is averaged. Here, $l$ is chosen to 5. A higher $l$ would ensure a flatter drop from the beginning over a longer length. The lowest target utility value of the human is in the time from $t=\SI{27}{\second}$ to $t=\SI{33}{\second}$ with a value of $U_\text{target}\approx 0.5$, while $\hat{x}_{\text{stubb,H},\omega}\approx 0.5$. After the human no longer communicates stubbornness from $t=\SI{33}{\second}$, the target utility of the automation increases again and it is the automation that determines the desire to move.

Fig. \ref{fig:sim_results} d) shows the estimated desired movement $\hat{\omega}_\text{H}$ of the human (yellow curve), which is also estimated from the curve for $F_\text{Int,y}$ according to \eqref{eq:omega_estimation_human} and \eqref{eq:movement_estimation_human}. The slight fluctuations of $\hat{\omega}_\text{H}$ result from the actually executed $\omega_\text{A}$, which is added to $\Delta\omega$ according to \eqref{eq:movement_estimation_human} and changes with each iteration depending on the negotiated angular velocity. The blue curve shows the desired movement of the automation $\omega_\text{A}^*$, which wants to return to its global path. The orange dashed curve shows the angular velocity $\omega_\text{A}$, which results from the agreement between the two agents. It can be seen that after the leaps in force and thus increasing stubbornness $\hat{x}_{\text{stubb,H},\omega}$ of the human, the automation does not stick to its angular velocity $\omega_\text{A}^*$, but takes the human's request into account in the time from $t=\SI{19}{\second}$ to $t=\SI{33}{\second}$ and accommodates it. It can also be seen that after the human's interaction force has decayed at $t=\SI{33}{\second}$ after two adjustment steps, the executed angular velocity $\omega_\text{A}$ from $t\approx\SI{34.4}{\second}$ corresponds completely to the automation's desired movement. From this time the estimated movement request of the human $\hat{\omega}_\text{H}$ corresponds to the same course as $\omega_\text{A}$, but shifted by one negotiation step. This results from the estimation model \eqref{eq:omega_estimation_human} and \eqref{eq:movement_estimation_human}, according to which the human's movement request results from the currently executed movement $\boldsymbol{p}_\text{A}$ plus a $\Delta \boldsymbol{p}$. Since the human no longer exerts any force from $t=\SI{33}{\second}$, it applies $\Delta \boldsymbol{p}=\boldsymbol{0}$ and corresponds to the currently executed $\boldsymbol{p}$.

Fig. \ref{fig:sim_results} e) shows the coordinate $y(t)$, which according to the system model \eqref{eq:system-model} results from the constant longitudinal velocity of $v=\SI{0.8}{\meter\per\second}$ and the angular velocity $\omega_\text{A}$ resulting from the agreement. It can be seen that the entire system performs an evasive movement in positive $y$-direction in response to the human force input up to a maximum of $y\approx 0.52$ and then returns to the global path of automation. 
\section{DISCUSSION} \label{sec:discussion}
The simulation results from Section \ref{sec:simulation_results} show that human stubbornness can be reliably estimated quantitatively from human input. In addition, the automation reacts to increasing human stubbornness with compliant behavior, as required by reciprocal tit-for-tat behavior. The question that arises from the simulation results is whether the estimation model for the desire to change movement according to \eqref{eq:omega_estimation_human} and \eqref{eq:movement_estimation_human} and the model for estimating human stubbornness according to \eqref{eq:stubbornness_omega} correctly reflect human behavior. Another question of interest here is the extent to which the desire to change movement $\Delta \boldsymbol{p}$ and the stubbornness of the human $\boldsymbol{x}_\text{stubb,H}$ can be read from one and the same variable $\boldsymbol{F}_\text{Int}$. This question is subject of further investigation. However, the aim of this paper was not to validate the models, i.e. to determine whether the models \eqref{eq:omega_estimation_human}, \eqref{eq:movement_estimation_human} and \eqref{eq:stubbornness_omega} are optimal representations for the movement request of the human and his stubbornness in relation to an error measure with the true values. The contribution is rather the extension of the framework so that human stubbornness can be converted into a general parameter $\boldsymbol{x}_\text{stubb,H}$ and be read off. This allows the automation to react to the human behavior in a targeted manner. It can be seen that an opposite behavior for automation could be determined from the estimated human stubbornness $\hat{\boldsymbol{x}}_\text{stubb,H}$: as $\hat{\boldsymbol{x}}_\text{stubb,H}$ increases, $U_\text{target}$ decreases. This represents the $U_\text{target}(\boldsymbol{x}_\text{stubb})$ sought in the problem statement.
%
\enlargethispage{-5cm}
\section{CONCLUSION} \label{sec:conclusion}
In this paper, an existing framework for cooperative trajectory planning between a human and a automation from \cite{schneider_negotiation-based_2022} was extended to include human stubbornness. This represents a behavioral parameter of the human when agreeing on a joint trajectory, which describes human's behavior in terms of sticking to its original desire to move. The reciprocal tit-for-tat behavior presented in \cite{schneider_negotiation-based_2022} could be successfully reproduced with the model $U_\text{target}(\boldsymbol{x}_\text{stubb})$ introduced here. In the following, a study is pending to validate and possibly refine the models used to estimate the human stubbornness and the desire to change movement.

\addtolength{\textheight}{-12cm}   





\bibliographystyle{IEEEtran}

\end{document}